\pdfoutput=1

\documentclass[11pt]{article}

\usepackage{emnlp2021}

\usepackage{times}
\usepackage{latexsym}

\usepackage[T1]{fontenc}

\usepackage[utf8]{inputenc}

\usepackage{microtype}
\usepackage{hyperref}
\usepackage{url}
\usepackage{times}
\usepackage{latexsym}
\usepackage{graphicx}
\usepackage{amsmath}
\usepackage{flexisym}
\usepackage{multirow}
\usepackage{array}
\usepackage{float}

\title{Transfer Learning in Conversational Analysis through Reusing Preprocessing Data as Supervisors}

\author{Joshua Yee Kim \\
  University of Sydney \\
  \texttt{josh.kim@sydney.edu.au} \\\And
  Tongliang Liu \\
  University of Sydney \\
  \texttt{tongliang.liu@sydney.edu.au} \\ \\\AND
  Kalina Yacef \\
  University of Sydney \\
  \texttt{kalina.yacef@sydney.edu.au} \\}

\begin{document}
\maketitle
\begin{abstract}
Conversational analysis systems are trained using noisy human labels and often require heavy preprocessing during multi-modal feature extraction. Using noisy labels in single-task learning increases the risk of over-fitting. Auxiliary tasks could improve the performance of the primary task learning during the same training -- this approach sits in the intersection of transfer learning and multi-task learning (MTL). In this paper, we explore how the preprocessed data used for feature engineering can be re-used as auxiliary tasks, thereby promoting the productive use of data. Our main contributions are: (1) the identification of sixteen beneficially auxiliary tasks, (2) studying the method of distributing learning capacity between the primary and auxiliary tasks, and (3) studying the relative supervision hierarchy between the primary and auxiliary tasks. Extensive experiments on IEMOCAP and SEMAINE data validate the improvements over single-task approaches, and suggest that it may generalize across multiple primary tasks.
\end{abstract}

\section{Introduction}
The sharp increase in uses of video-conferencing creates both a need and an opportunity to better understand these conversations \citep{kim2019review}. In post-event applications, analyzing conversations can give feedback to improve communication skills \citep{hoque2013mach, naim2015automated}. In real-time applications, such systems can be useful in legal trials, public speaking, e-health services, and more \citep{poria2019emotion, tanveer2015rhema}.

Analyzing conversations requires both human expertise and a lot of time. However, to build automated analysis systems, analysts often require a training set annotated by humans \citep{poria2019emotion}. The annotation process is costly, thereby limiting the amount of labeled data. Moreover, third-party annotations on emotions are often noisy. Deep networks coupled with limited noisy labeled data increases the chance of overfitting \citep{james2013introduction, zhang2016understanding}. Could data be used more productively?

From the perspective of feature engineering to analyze video-conferences, analysts often employ pre-built libraries  \citep{baltruvsaitis2016openface, vokaturi2019} to extract multimodal features as inputs to training. This preprocessing phase is often computationally heavy, and the resulting features are only used as inputs. In this paper, we investigate how the preprocessed data can be re-used as auxiliary tasks which provide inductive bias through multiple noisy supervision \citep{caruana1997multitask, lipton2015learning, ghosn1997multi} and consequently,
promoting a more productive use of data. Specifically, our main contributions are  (1) the identification of beneficially auxiliary tasks, (2) studying the method of distributing learning capacity between the primary and auxiliary tasks, and (3) studying the relative supervision hierarchy between the primary and auxiliary tasks.  We demonstrate the value of our approach through predicting emotions on two publicly available datasets, IEMOCAP \citep{busso2008iemocap} and SEMAINE \citep{mckeown2011semaine}.

\section{Related Works and Hypotheses}
\label{section:related_works}

Multitask learning has a long history in machine learning \citep{caruana1997multitask}. In this paper, we focus on transfer learning within MTL, a less commonly discussed subfield within MTL \citep{mordan2018revisiting}. We are concerned with the performance on one (primary) task -- the sole motivation of adding auxiliary tasks is to improve the primary task performance.

In recent years, this approach has been gaining attention in computer vision \citep{yoo2018deep, fariha2016automatic, yang2018multitask, mordan2018revisiting, sadoughi2018expressive}, speech recognition \citep{krishna2018hierarchical, chen2015multitask, tao2020end, bell2016multitask, chen2014joint}, and natural language processing (NLP) \citep{arora2019multitask, yousif2018citation, zalmout2019adversarial, yang2019information, du2017novel}. The drawback of adding multiple tasks increases the risk of negative transfer \citep{torrey2010transfer, lee2016asymmetric, lee2018deep,liu2019loss, simm2014tree}, which leads to many design considerations. Three of such considerations are, identifying (a) what tasks are beneficial, (b) how much of the model parameters to share between the primary and auxiliary tasks, and (c) whether we should prioritize primary supervision by giving it a higher hierarchy than the auxiliary supervision. 

In contrast with previous MTL works, our approach (a) identifies sixteen beneficially auxiliary targets, (b) dedicates a primary-specific branch within the network, and (c) investigates the efficacy and generalization of prioritizing primary supervision across eight primary tasks.

Since our input representation is fully text-based, we dive deeper into MTL model architecture designs in NLP, \citet{sogaard2016deep} found that lower-level tasks like part-of-speech tagging, are better kept at the lower layers, enabling the higher-level tasks like Combinatory Categorical Grammar tagging to use these lower-level representations. In our approach, our model hierarchy is not based on the difficulty of the tasks, but more simply, we prioritize the primary task. Regarding identifying auxiliary supervisors in NLP, existing works have included tagging the input text \citep{zalmout2019adversarial,yang2019information, sogaard2016deep}. Text classification with auxiliary supervisors have included research article classification \citep{du2017novel, yousif2018citation}, and tweet classification \citep{arora2019multitask}. 

Multimodal analysis of conversations has been gaining attention in deep learning research \citep{poria2019emotion}. The methods in the recent three years have been intelligently fusing numeric vectors from the text, audio, and video modalities before feeding it to downstream layers. This approach is seen in 
MFN \citep{zadeh2018memory},
MARN \citep{zadeh2018multi},
CMN \citep{hazarika2018conversational}, 
ICON \citep{hazarika2018icon},
DialogueRNN \citep{majumder2019dialoguernn},
and M3ER \citep{mittal2020m3er}.  Our approach is different in two ways.
(1) Our audio and video information is encoded within text before feeding only the text as input. Having only text as input has the benefits of interpretability, and the ability to present the conversational analysis on paper \citep{kim2019detecting}, similar to how the linguistics community performs manual conversational analysis using the Jefferson transcription system \citep{jefferson2004glossary}, where the transcripts are marked up with symbols indicating how the speech was articulated. 
(2) Instead of using the audio and video information as only inputs, we demonstrate how to use multimodal information in both input and as auxiliary supervisors.

\textbf{Hypothesis H1: The introduced set of auxiliary supervision features improves primary task performance.} We introduce and motivate the full set of sixteen auxiliary supervisions, all based on existing literature: these are grouped into four families, each with four auxiliary targets. The four families are (1) facial action units, (2) prosody, (3) historical labels, and (4) future labels: 
\newline (1) Facial action units, from the facial action coding system identifies universal facial expressions of emotions \citep{ekman1997face}. Particularly, AU 05 (upper lid raiser), 17 (chin raiser), 20 (lip stretcher), 25 (lips part) have been shown to be useful in detecting depression \citep{yang2016decision, kim2019detecting} and rapport-building \citep{kim2021monah}. \newline  
(2) Prosody, the tone of voice -- happiness, sadness, anger, and fear -- can project warmth and attitudes \citep{hall2009observer}, and has been used as inputs in emotions detection \citep{garcia2017emotion}. \newline
(3 and 4) Using features at different historical time-points is a common practice in statistical learning, especially in time-series modelling \citep{christ2018time}. Lastly, predicting future labels as auxiliary tasks can help in learning \citep{caruana1996using, cooper2005predicting, trinh2018learning, zhu2020vision, shen2020auxiliary}. We propose using historical and future (up to four talkturns ago or later) target labels as auxiliary targets.

\begin{figure*}[ht]
    \includegraphics[width=1.0\textwidth]{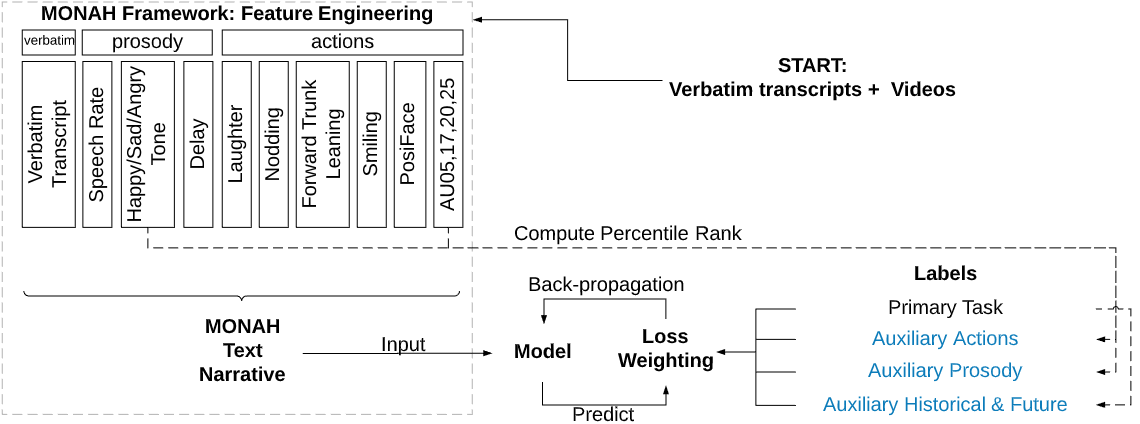}
    \caption{Reusing components (dotted lines) of the feature engineering process as auxiliary targets (in blue). The MONAH framework \citep{kim2021monah} is introduced in section \ref{ssect:input}}
    \label{fig:reuse}
\end{figure*}

Given that we are extracting actions and prosody families as inputs, we propose to explore whether they can be reused as supervisors (see Fig. \ref{fig:reuse}). Our hypothesis \textbf{H1} is that re-using them as auxiliary supervision improves primary task performance. This is related to using hints in the existing MTL literature, where the auxiliary tasks promote the learning of the feature \citep{cheng2015open, yu2016learning}.

\textbf{Hypothesis H2: When the primary branch is given maximum learning capacity, it would not be outperformed by models with primary branch having less than the maximum learning capacity.} Deeper models with higher learning capacity produce better results \citep{huang2019gpipe, nakkiran2019deep, menegola2017knowledge, blumberg2018deeper, romero2014fitnets}. Also, since the auxiliary branch is shared with the primary supervision, the auxiliary capacity should be limited to improve primary task performance \citep{wu2020understanding} because limiting the auxiliary capacity will force the branch to learn common knowledge (instead of auxiliary specific knowledge) across the auxiliary tasks \citep{arpit2017closer}. Therefore, given a fixed learning capacity budget, our hypothesis \textbf{H2} implies that we should allocate the maximum learning capacity to the primary branch because we care only about the primary task performance.

\textbf{Hypothesis H3: Auxiliary supervision at the lower hierarchy yields better primary task performance as compared to flat-MTL.} Having the auxiliary tasks at the same supervisory level as the primary task is inherently sub-optimal because we care only about the performance of the primary task \citep{mordan2018revisiting}. Information at the lower hierarchy learns basic structures that are easy to transfer, whilst upper hierarchy learns more semantic information that is less transferable \citep{zeiler2014visualizing}. Therefore, we propose that the auxiliary supervision to be at the lower hierarchy than the primary supervision.

\section{Model Architecture}
\subsection{Flat-MTL Hierarchical Attention Model}
\label{ssect:flat_mtl_han}
We start with an introduction of the Hierarchical Attention Model (HAN) \citep{yang2016hierarchical}. We chose HAN because of its easy interpretability as it only uses single-head attention layers. There are four parts to the HAN model, (1) text input, (2) word encoder, (3) talkturn encoder, and (4) the predictor. In our application, we perform our predictions at the talkturn-level for both IEMOCAP and SEMAINE. For notation, let $s_i$ represent the \textit{i}-th talkturn and $w_{it}$ represent the \textit{t}-th word in the \textit{i}-th talkturn. Each single talkturn can contain up to \textit{T} words, and each input talkturn can contain up to \textit{L} past talkturns to give content context (see section \ref{ssect:input}).

Given a talkturn of words, we first convert the words into vectors through an embedding matrix $W_e$, and the word selection one-hot vector, $w_{it}$. The word encoder comprises of bidirectional GRUs \citep{bahdanau2014neural} and a single head attention to aggregate word embeddings into talkturn embeddings. Given the vectors $x_{it}$, the bidirectional GRU reads the words from left to right as well as from right to left (as indicated by the direction of the GRU arrows) and concatenates the two hidden states together to form $h_{it}$. We then aggregate the hidden states into one talkturn embedding through the attention mechanism. $u_{it}$ is the hidden state from feeding $h_{it}$ into a one-layer perceptron (with weights $W_{w}$ and biases $b_{w}$). The attention weight ($\alpha_{it}$) given to $u_{it}$ is the softmax normalized weight of the similarity between itself ($u_{it}$) and $u_w$, which are all randomly initialized and learnt jointly.

\begin{figure*}[ht]
    \includegraphics[width=1.0\textwidth]{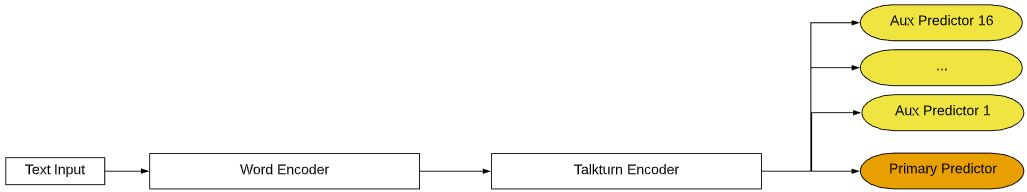}
    \caption{Forward pass of the Flat-MTL HAN architecture. Auxiliary tasks (yellow) are added at the same level of the primary task (orange).}
    \label{fig:han}
\end{figure*}

\begin{figure*}[ht]
    \includegraphics[width=1.0\textwidth]{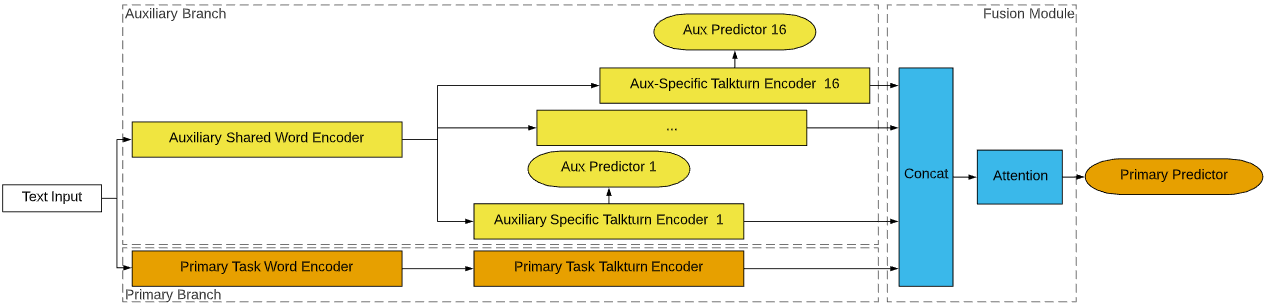}
    \caption{Forward pass of the HAN-ROCK architecture. There is a primary branch (orange) where auxiliary supervision (yellow) can not influence. The fusion module (blue) aggregates the talkturn embeddings from all tasks into one.}
    \label{fig:rock}
\end{figure*}

\begin{gather*} 
x_{it} = W_e w_{it}, t \in [1,T]. \\
\overrightarrow{h}_{it} = \overrightarrow{GRU}(x_{it}), t \in [1,T]. \\
\overleftarrow{h}_{it} = \overleftarrow{GRU}(x_{it}), t \in [T,1]. \\
{h}_{it} = (\overrightarrow{h}_{it}, \overleftarrow{h}_{it}). \\
u_{it} = relu(W_w h_{it} + b_{w}).
\end{gather*}
\begin{gather*} 
s_i = \Sigma_t \alpha_{it} u_{it}. \\
\alpha_{it} = \frac{exp(u_{it}^{\top} u_w)}{\Sigma_t exp(u_{it}^{\top} u_w)}.
\end{gather*}

With the current and past talkturn embeddings (content context, to discuss in section \ref{ssect:input}), the talkturn encoder aggregates them into a single talkturn representation ($v$) in a similar fashion, as shown below.

\begin{gather*}
\overrightarrow{h}_{i} = \overrightarrow{GRU}(s_{i}), i \in [1,L]. \\
\overleftarrow{h}_{i} = \overleftarrow{GRU}(s_{i}), i \in [L,1]. \\
{h}_{i} = (\overrightarrow{h}_{i}, \overleftarrow{h}_{i}). \\
u_{i} = relu(W_s h_{i} + b_{s}). \\
\alpha_{i} = \frac{exp(u_{i}^{\top} u_s)}{\Sigma_i exp(u_{i}^{\top} u_s)}. \\
v = \Sigma_i \alpha_{i} u_{i}.
\end{gather*}

The simplest way of adding the sixteen auxiliary task predictors would be to append them to where the primary task predictor is, as illustrated in Fig. \ref{fig:han}. That way, all predictors use the same representation $v$. We refer to this architecture as flat-MTL, but we are unable to test \textbf{H2} and \textbf{H3} using this architecture. Therefore, we introduce HAN-ROCK next. 

\subsection{HAN-ROCK}
\label{ssect:han_rock}
We adapted\footnote{Implementation will be available at GitHub; Please see attached supplementary material during the review phase.} the ROCK architecture \citep{mordan2018revisiting} which was built for Convolutional Neural Networks \citep{lecun1995convolutional} found in ResNet-SSD \citep{he2016deep, liu2016ssd} to suit GRUs \citep{bahdanau2014neural} found in HAN \citep{yang2016hierarchical} (see Fig. \ref{fig:rock}). To study \textbf{H3}, we bring the auxiliary task predictors forward (see Fig. \ref{fig:rock}), so that the back-propagation from the primary supervision is able to temper the back-propagation from the auxiliary supervision but not vice-versa. This also sets us up to study \textbf{H2}. Each of the auxiliary tasks has its own talkturn encoder but shares one word encoder in the auxiliary branch (to keep the network small). Subscript $a$ indicates whether the word encoder is for the primary or auxiliary branch:
\begin{gather*}
x_{it} = W_e w_{it}, t \in [1,T] \\
\overrightarrow{h}_{ait} = \overrightarrow{GRU_a}(x_{it}), t \in [1,T], a \in \{pri, aux\} \\
\overleftarrow{h}_{ait} = \overleftarrow{GRU_a}(x_{it}), t \in [T,1], a \in \{pri, aux\} \\
{h}_{ait} = (\overrightarrow{h}_{ait}, \overleftarrow{h}_{ait}) \\
u_{ait} = relu(W_{aw} h_{ait} + b_{aw}) \\
\alpha_{ait} = \frac{exp(u_{ait}^{\top} u_{aw})}{\Sigma_t exp(u_{ait}^{\top} u_{aw})} \\
s_{ai} = \Sigma_{t} \alpha_{ait} u_{ait}
\end{gather*}

\begin{figure*}[ht]
    \centering
    \includegraphics[width=0.75\textwidth]{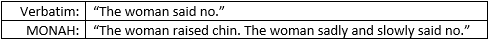}
    \caption{Example of a MONAH transcript.}
    \label{fig:verbatim}
\end{figure*}

Each task has its own talkturn encoder. Subscript $b$ indicates which of the seventeen tasks -- the primary talkturn task or one of the sixteen auxiliary tasks -- is the talkturn encoder is dedicated to: 

\begin{gather*}
\overrightarrow{h}_{abi} = \overrightarrow{GRU_{ab}}(s_{ai}), i \in [1,L], a \in \{pri, aux\}, \\ 
b \in \{pri, aux1, aux2, ..., aux16 \}  \\
\overleftarrow{h}_{abi} = \overleftarrow{GRU_{ab}}(s_{ai}), i \in [L,1], a \in \{pri, aux\}, \\ 
b \in \{pri, aux1, aux2, ..., aux16 \}   \\
{h}_{abi} = (\overrightarrow{h}_{abi}, \overleftarrow{h}_{abi}) \\
u_{abi} = relu(W_b h_{abi} + b_{b}) \\
\alpha_{abi} = \frac{exp(u_{abi}^{\top} u_b)}{\Sigma_i exp(u_{abi}^{\top} u_b)} \\
v_{ab} = \Sigma_i \alpha_{abi} u_{abi}
\end{gather*}

The seventeen talkturn embeddings ($v_{ab}$) goes through a concatenation, then the single head attention, aggregating talkturn embeddings across seventeen tasks into one talkturn embedding for the primary task predictor. Subscript $c$ pertains to the fusion module.
\begin{gather*}
\text{concatenation: } v_c = (v_{ab}), a \in \{pri, aux\}, \\
b \in \{pri, aux1, aux2, ..., aux16 \}  \\
\text{attention: }
\alpha_{c} = \frac{exp(v_c^{\top} u_c)}{\Sigma_c exp(v_{c}^{\top} u_c)} \\
\text{overall primary talkturn vector: }
v = \Sigma_c \alpha_{c} v_{c}
\end{gather*}

\section{Experiments}
\subsection{Data and Primary Tasks}

We validate our approach using two datasets with a total of eight primary tasks: the IEMOCAP \citep{busso2008iemocap} and the SEMAINE \citep{mckeown2011semaine} datasets. Both datasets are used in multimodal emotions detection research \citep{poria2019emotion}. We divided the datasets into train, development, and test sets in an approximate 60/20/20 ratio such that the sets do not share any speaker (Appendix \ref{appendix:partition} details the splits).

The target labels of the eight primary tasks are all at the talkturn-level. The four primary tasks of IEMOCAP consists of the four-class emotions classification (angry, happy, neutral, sad), and three regression problems -- valence (1-negative, 5-positive), activation (1-calm, 5-excited), and dominance (1-weak, 5-strong). The four-class emotions classification target is common for IEMOCAP \citep{latif2020multi, xia2015multi, li2019improved, hazarika2018conversational, mittal2020m3er}. For SEMAINE, there are four regression problems -- activation, intensity, power, valence. We use two standard evaluation metrics, mean absolute error (MAE), and 4-class weighted mean classification accuracy, MA(4).

\subsection{Input}
\label{ssect:input}
Multimodal feature extraction is computed using the MONAH framework \citep{kim2021monah}. This framework uses a variety of pre-trained models to extract nine multimodal features, associated with the prosody of the speech and the actions of the speaker, and weaves them into a multimodal text narrative. We refer the reader to \citet{kim2021monah} for the details and efficacy of the MONAH framework. The benefit of the created narrative is that it describes what is said together with how it is said for each talkturn, giving richer nonverbal context to the talkturn (see Fig. \ref{fig:verbatim} for an example). Being fully text-based means that the analysis product can be printed out on paper, without the need for speakers nor monitors to replay the conversation on a computer.

In addition to nonverbal context, we concatenated a variable number of preceding talkturns to the current talkturn as content context. Content context has been proven to be useful in CMN \citep{hazarika2018conversational} and ICON \citep{hazarika2018icon}, DialogueRNN \citep{majumder2019dialoguernn}. The content-context size is tuned as a hyperparameter. The resulting multimodal text narrative, consisting of both nonverbal and context context, is used as the sole input to the model.

\subsection{Auxiliary Targets}

We first clarify the method of extraction for the auxiliary families. The OpenFace algorithm \citep{baltruvsaitis2016openface} is used to extract the four continuous facial action units (AU) -- AU 05, 17, 20, 25. The Vokaturi algorithm \citep{vokaturi2019} is used to extract the four continuous dimensions in the tone of voice -- happiness, sadness, anger, and fear. As for historical and future features, we simply look up the target label for the past four talkturns and future four talkturns. Any label that is not available (for example, the label four talkturns ago is not available for the third talkturn) is substituted with the next nearest non-missing label.

All auxiliary targets that reused the input features (actions and prosody) are converted into a percentile rank that has the range [0,1] using the values from the train partition. This is a subtle but note-worthy transformation. When reusing an input as an auxiliary target, it would be trivial if the input can easily predict the target. For example, given the following MONAH transcript as input, ``The woman sadly and slowly said no." It would be trivial to use a binary (quantized) auxiliary target of ``was the tone sad?" because we would only be training the model to look for the word ``sadly". However, if the auxiliary target is a percentile rank (less quantized) of the sadness in tone, then the presence of the word ``sadly" increases the predicted rank, but the model could still use the rest of the nonverbal cues (``slowly") and what is being said (``no") to predict the degree of sadness. That way, representations learnt for the auxiliary tasks uses more of the input.

Percentile rank also has the convenient property of having the range [0,1]. We scaled the percentile ranks so that they all have the same range as the primary task (see appendix \ref{appendix:scalingaux} for transformation details). This ensures that if we assigned equal loss weights to all tasks, the contribution of every task is of the same order of magnitude \citep{gong2019comparison, hassani2019unsupervised, sener2018multi}. 

\begin{table*}[ht]
  \centering
  \caption{\textbf{H1} Results. *: the model performance has a statistically significant difference with the baseline model. a: action, p: prosody, h: historical labels, f: future labels.}
  \label{tab:h1}
  \begin{tabular}{|>{\centering\arraybackslash}m{1.50cm}
                |>{\centering\arraybackslash}m{1.00cm}
                |>{\centering\arraybackslash}m{1.00cm}
                |>{\centering\arraybackslash}m{1.00cm}
                |>{\centering\arraybackslash}m{1.00cm}
                |>{\centering\arraybackslash}m{1.00cm}
                |>{\centering\arraybackslash}m{1.00cm}
                |>{\centering\arraybackslash}m{1.00cm}
                |>{\centering\arraybackslash}m{1.00cm}|}
\hline 
        { }     &   \multicolumn{4}{c|}{\textbf{IEMOCAP}} &   \multicolumn{4}{c|}{\textbf{SEMAINE}} \\
\hline
            {\textbf{Aux. Target}}     &   {\textbf{Classif. \newline MA(4)} }
        &   {\textbf{Val. \newline MAE}}   &   {\textbf{Act. \newline MAE}}
        &   {\textbf{Dom. \newline MAE}}  &   {\textbf{Act. \newline MAE}}
        &   {\textbf{Int. \newline MAE}}  &   {\textbf{Pow. \newline MAE}}
        &   {\textbf{Val. \newline MAE}}\\
\hline
            {None (Baseline)}     &   {0.625}
        &   {0.527}   &   {0.518}
        &   {0.667}  &   {0.194}
        &   {0.238}  &   {0.170}
        &   {0.177}    \\
\hline
            {ap}     &   {0.715*}
        &   {0.538}   &   {0.507}
        &   {0.600*} &   {0.184*}
        &   {0.218*}  &   {0.167*}
        &   {0.178}     \\
\hline
            {aphf}     &   {0.706*}
        &   {0.497*}   &   {0.504}
        &   {0.587*}&   {0.187}
        &   {0.231}  &   {0.165*}
        &   {0.169}     \\
\hline
\end{tabular}
\end{table*}

\begin{table*}[ht]
  \centering
  \caption{\textbf{H2} Results. *: the model performance has a statistically significant difference with the baseline model (\textit{P} = 256). ^: Assigning 1 GRU to the auxiliary task talkturn encoder yields a statistically significant difference with assigning 0 GRU.}
  \label{tab:h2}
  \begin{tabular}{|>{\centering\arraybackslash}m{1.50cm}
                |>{\centering\arraybackslash}m{1.00cm}
                |>{\centering\arraybackslash}m{1.00cm}
                |>{\centering\arraybackslash}m{1.00cm}
                |>{\centering\arraybackslash}m{1.00cm}
                |>{\centering\arraybackslash}m{1.00cm}
                |>{\centering\arraybackslash}m{1.00cm}
                |>{\centering\arraybackslash}m{1.00cm}
                |>{\centering\arraybackslash}m{1.00cm}|}
\hline 
        { }     &   \multicolumn{4}{c|}{\textbf{IEMOCAP}} &   \multicolumn{4}{c|}{\textbf{SEMAINE}} \\
\hline
            {\textbf{Pri. GRU}}     &   {\textbf{Classif. \newline MA(4)} }
        &   {\textbf{Val. \newline MAE}}   &   {\textbf{Act. \newline MAE}}
        &   {\textbf{Dom. \newline MAE}}  &   {\textbf{Act. \newline MAE}}
        &   {\textbf{Int. \newline MAE}}  &   {\textbf{Pow. \newline MAE}}
        &   {\textbf{Val. \newline MAE}}\\
\hline
            {256 (Baseline)}     &   {0.715^}
        &   {0.538}   &   {0.507}
        &   {0.587^}  &   {0.187}
        &   {0.231}  &   {0.165^}
        &   {0.169^}    \\
\hline
            {192}     &   {0.736}
        &   {0.509}   &   {0.518}
        &   {0.604} &   {0.187}
        &   {0.228}  &   {0.165}
        &   {0.184*}     \\
\hline
            {128}     &   {0.711}
        &   {0.537}   &   {0.512}
        &   {0.597}   &   {0.189}
        &   {0.216}  &   {0.176*}
        &   {0.196*}     \\
\hline
            {64}     &   {0.687}
        &   {0.540}   &   {0.507}
        &   {0.593}&   {0.191}
        &   {0.234}  &   {0.167}
        &   {0.192*}     \\
\hline
            {1}     &   {0.656}
        &   {0.554}   &   {0.509}
        &   {0.599} &   {0.190}
        &   {0.229}  &   {0.168*}
        &   {0.191*}     \\
\hline
\end{tabular}
\end{table*}

\subsection{Models, training, and hyperparameters tuning}
\label{ssect:models}

The overall loss is calculated as the weighted average across all seventeen tasks: (1) we picked a random weight for the primary task from the range [0.50, 0.99]; this ensures that the primary task has the majority weight. (2) For the remaining weights (1 - primary weight), we allocated them to the sixteen auxiliary tasks by: (a) random, (b) linearly-normalized mutual information, or (c) softmax-normalized mutual information. (a) is self-explanatory. As for (b) and (c), mutual information has been shown that it is the best predictor -- compared to entropy and conditional entropy -- of whether the auxiliary task would be helpful \citep{bjerva2017will}. We computed the mutual information (vector $m$) of each auxiliary variable with the primary target variable \citep{kraskov2004estimating, ross2014mutual} using scikit-learn \citep{scikit-learn}. Then, we linearly-normalized or softmax-normalized $m$ to sum up to 1. Finally, we multiplied the normalized $m$ with the remaining weights from (2); this ensures that the primary weight and the sixteen auxiliary weight sum up to one. (a), (b), and (c) have ten trials each during hyper-parameters tuning.

Two variants of the HAN architectures are used (Fig \ref{fig:han} and \ref{fig:rock}). For hypotheses testing, we bootstrapped confidence intervals (appendix \ref{appendix:bootstrap}).

\section{Results and Discussion}
The key takeaways are: (\textbf{H1}) The introduced set of auxiliary supervision improves primary task performance significantly in six of the eight primary tasks.
(\textbf{H2}) Maximum learning capacity should be given to the primary branch as a default. (\textbf{H3}) HAN-ROCK is unlikely (in one of the eight tasks) to degrade primary task performance significantly, and sometimes significantly improves it (in four of the eight tasks).

(\textbf{H1}): To test \textbf{H1} (whether the introduced set of auxiliary supervision improves primary task performance), we first train the model with all sixteen auxiliary targets (from families: actions, prosody, historical, and future). Then, to differentiate the effect from the historical and future supervision, we set the loss weights from historical and future targets to be zero; effectively, there is only supervision from eight auxiliary targets (actions and prosody). Lastly, for the baseline model (no auxiliary supervision), we set the loss weights from all sixteen auxiliary targets to zero.

\begin{table*}[ht]
  \centering
  \caption{\textbf{H3} Results. *: the model performance has a significant difference with the baseline.}
  \label{tab:h3}  
  \begin{tabular}{|>{\centering\arraybackslash}m{1.50cm}
                |>{\centering\arraybackslash}m{1.00cm}
                |>{\centering\arraybackslash}m{1.00cm}
                |>{\centering\arraybackslash}m{1.00cm}
                |>{\centering\arraybackslash}m{1.00cm}
                |>{\centering\arraybackslash}m{1.00cm}
                |>{\centering\arraybackslash}m{1.00cm}
                |>{\centering\arraybackslash}m{1.00cm}
                |>{\centering\arraybackslash}m{1.00cm}|}
\hline 
        { }     &   \multicolumn{4}{c|}{\textbf{IEMOCAP}} &   \multicolumn{4}{c|}{\textbf{SEMAINE}} \\
\hline
            {\textbf{Hierarchy}}     &   {\textbf{Classif. \newline MA(4)} }
        &   {\textbf{Val. \newline MAE}}   &   {\textbf{Act. \newline MAE}}
        &   {\textbf{Dom. \newline MAE}}  &   {\textbf{Act. \newline MAE}}
        &   {\textbf{Int. \newline MAE}}  &   {\textbf{Pow. \newline MAE}}
        &   {\textbf{Val. \newline MAE}}\\
\hline
            {Flat (Baseline)}     &   {0.699}
        &   {0.520}   &   {0.526}
        &   {0.606}  &   {0.183}
        &   {0.230}  &   {0.164}
        &   {0.185}    \\
\hline
            {HAN-ROCK}     &   {0.715}
        &   {0.538}   &   {0.507*}
        &   {0.600*} &   {0.184}
        &   {0.218*}  &   {0.167*}
        &   {0.178*}     \\
\hline
\end{tabular}
\end{table*}

\begin{table*}[ht]
  \centering
  \caption{Class-wise classification F1 score on IEMOCAP. Baseline (challenger) refers to HAN-Rock architecture under the three hypotheses. *: the challenger performance has a statistically significant difference with the baseline model.}
  \label{tab:low_resource}
  \begin{tabular}{|>{\centering\arraybackslash}m{1.00cm}
                |>{\centering\arraybackslash}m{1.00cm}
                |>{\centering\arraybackslash}m{1.20cm}
                |>{\centering\arraybackslash}m{1.70cm}
                |>{\centering\arraybackslash}m{1.20cm}
                |>{\centering\arraybackslash}m{1.7cm}
                |>{\centering\arraybackslash}m{1.20cm}
                |>{\centering\arraybackslash}m{1.7cm}
                |>{\centering\arraybackslash}m{0.8cm}
                |}
                
\hline 
        \multicolumn{2}{|c|}{\textbf{Distribution}}
        &   \multicolumn{2}{c|}{\textbf{H1}}
        &   \multicolumn{2}{c|}{\textbf{H2}}
        &   \multicolumn{2}{c|}{\textbf{H3}}
        &   {\textbf{SoTA}}
        \\
\hline
            {\textbf{Label}}     &   {\textbf{Count}}
        &   {\textbf{Baseline \newline Aux target: None}}   
        &   {\textbf{Challenger \newline Aux target: aphf}} 
        &   {\textbf{Baseline \newline Pri GRU: 256}}   
        &   {\textbf{Challenger \newline Pri GRU: 1}} 
        &   {\textbf{Baseline \newline Hier-archy: Flat}}   
        &   {\textbf{Challenger \newline Hier-archy: HAN ROCK}} 
        &   {\textbf{M3-ER}}\\
\hline
            {Sad}     &{1,084} &{0.573} & {0.689*} &{0.699} 
                      &{0.591*} &{0.674} &{0.704*} &{0.775}\\
\hline
            {Anger}   &{1,103} &{0.531} & {0.683} &{0.752} 
                      &{0.672*} &{0.657} &{0.720*} &{0.862}\\
\hline
            {Happy}   &{1,636} &{0.772} & {0.784} &{0.776} 
                      &{0.754*} &{0.804} &{0.806} &{0.862}\\
\hline
            {Neutral} &{1,708} &{0.664} & {0.636} &{0.688} 
                      &{0.627*} &{0.645} &{0.631} &{0.745}\\
\hline
\end{tabular}


\end{table*}

Given auxiliary supervision, the model significantly outperforms the baseline of not having auxiliary supervision in six out of the eight primary tasks (Table \ref{tab:h1}). Comparing the baseline model with the model with two auxiliary target families, they significantly outperformed the baseline model in five out of eight primary tasks. The addition of two auxiliary target families (historical and future labels) sometimes significantly improved primary task performance (valence in IEMOCAP), but it also sometimes significantly made it worse (activation and intensity in SEMAINE). This shows that the value of auxiliary tasks, and the associated risk of negative transfer, depends on the auxiliary task.

(\textbf{H2}): To test \textbf{H2} (whether maximum learning capacity should be given to the primary branch), we let \textit{P} represent the number of GRU assigned to the primary talkturn encoder, and \textit{A} represent the number of GRU assigned to each of the sixteen auxiliary talkturn encoder. We constrained \textit{P} + \textit{A} to be equal to 257. During our experiments, we set \textit{P} to 1, 64, 128, 192, and 256. We set 256 as the baseline model because it is the maximum learning capacity we can give to the primary branch while giving 1 GRU ($=257-256$) to each of the sixteen auxiliary talkturn encoders.

In all primary tasks, the baseline model of assigning 256 GRUs to the primary branch is not significantly outperformed by models that assigned 1, 64, 128, 192 GRUs (Table \ref{tab:h2}). Generally, the performance decreased as the number of GRUs assigned to the primary talkturn encoder decreased from 256 to 1. We observed significantly worse performance in two out of eight tasks -- in power and valence in SEMAINE. Also, assigning 256 GRU to the primary talkturn encoders and 1 to each of the sixteen auxiliary talkturn encoders yields the smallest model\footnote{As opposed to assigning 1 GRU to the primary talkturn encoder and 256 GRU to each of the sixteen auxiliary encoder.}, and thus trains the fastest. Therefore, we recommend that the maximum capacity be given to the primary branch as a default. 

That said, the presence of an auxiliary branch is still important. The baseline of \textbf{H1} (no auxiliary supervision, Table \ref{tab:h1}) can be approximated\footnote{Same model architecture except that the loss weights of all auxiliary tasks are zero} as \textit{P}=$256 + 16 \times 1$, \textit{A}=0 . We compared the former to the baseline in Table \ref{tab:h2}, and found that four out of eight primary tasks have significant improvements by changing the number of talkturn encoders assigned to each auxiliary task from zero to \textit{one}.

(\textbf{H3}): To test \textbf{H3} (whether auxiliary supervision should be given a lower hierarchy), we compare the results from the flat-MTL HAN architecture (baseline) against the HAN-ROCK architecture (Table \ref{tab:h3}). Placing auxiliary supervision at the lower hierarchy significantly improves primary task performance in four out of eight tasks. In only one out of eight tasks (power in SEMAINE), auxiliary supervision significantly degrades primary task performance. Further improvements are possible through the fusion module with future research. 

\subsection{Class-wise Performance and SoTA}
Generally, we found that all hypotheses effects are stronger in lower resource labels (sad and anger, Table \ref{tab:low_resource}). We also present the performance of M3ER \citep{mittal2020m3er}, a previous state-of-the-art (SoTA) approach. We do not expect the performance of our text-only input to match the SoTA approach, which is confirmed in Table \ref{tab:low_resource}. By fusing numerical vectors from the three modalities prevalent in SoTA approaches \citep{zadeh2018memory, zadeh2018multi, hazarika2018conversational, hazarika2018icon, majumder2019dialoguernn, mittal2020m3er}, the inputs are of a much higher granularity as compared to our approach of describing the multimodal cues using discrete words. Although the text-based input is likely to constrain model performance, the multimodal transcription could be helpful for a human to analyze the conversation, we could also overlay the model perspective on the multimodal transcription to augment human analysis (see Appendix \ref{appendix:visualization}).

\section{Conclusion}
We proposed to re-use feature engineering pre-processing data as auxiliary tasks to improve performance and transfer learning. Three hypotheses were tested. The experimental results confirm \textbf{H1} -- Introducing our set of sixteen auxiliary supervisors resulted in better performance in most primary tasks. For \textbf{H2}, maximum learning capacity should be given to the primary branch. Lastly, for \textbf{H3}, placing the auxiliary supervision in a lower hierarchy is unlikely to hurt performance significantly, and it sometimes significantly improves performance. This is encouraging news for multi-modal conversational analysis systems as we have demonstrated how pre-processed data can be used \textit{twice} to improve performance, once as inputs, and again as auxiliary tasks. 

The first limitation of our paper is that the solutions are evaluated on eight tasks in the conversational analysis domain, and it is not clear if these would generalize outside of this domain. The second limitation is that we have evaluated on HAN, but not on other network architectures.

A challenge to be addressed is the apriori selection of the auxiliary targets. Future research could investigate targets selection, including how to use a much larger range of auxiliary targets, how to decide the optimum number of auxiliary targets, and whether it is possible to perform these automatically.

\newpage
\bibliographystyle{acl_natbib}
\bibliography{anthology,custom}

\newpage
\appendix
\section{Appendix}

\subsection{Dataset partitions}
\label{appendix:partition}
We detail the dataset partition in Table \ref{tab:partition} for aid reproducibility.

\begin{table*}[ht]
  \caption{Dataset partitions}
  \label{tab:partition}
  \centering
  \begin{tabular}{|>{\centering\arraybackslash}m{1.50cm}
                |>{\centering\arraybackslash}m{1.50cm}
                |>{\centering\arraybackslash}m{1.20cm}
                |>{\centering\arraybackslash}m{1.50cm}
                |>{\centering\arraybackslash}m{3.00cm}
                |>{\centering\arraybackslash}m{1.20cm}|}
\hline 
         \multicolumn{3}{|c|}{\textbf{IEMOCAP}} &   \multicolumn{3}{c|}{\textbf{SEMAINE}} \\
\hline
        {\textbf{Partition} } & {\textbf{Session}} & {\textbf{Count}} 
        & {\textbf{Partition} } & {\textbf{Session}} & {\textbf{Count}} 
        \\
\hline {Train} &   {Ses01F}   &   {861} &   {Train}  &   {2008.12.05.16.03.15} &   {424}  \\
\hline {Train} &   {Ses01M}   &   {958} &   {Train}  &   {2009.01.30.12.00.35} &   {404}  \\
\hline {Train} &   {Ses02F}   &   {889} &   {Train}  &   {2009.02.12.10.49.45} &   {686}  \\
\hline {Train} &   {Ses02M}   &   {922} &   {Train}  &   {2009.05.15.15.04.29} &   {436}  \\
\hline {Train} &   {Ses03F}   &   {958} &   {Train}  &   {2009.05.29.14.30.05} &   {760}  \\
\hline {Train} &   {Ses03M}   &   {1178} &   {Train}  &   {2009.05.22.15.17.45} &   {668}  \\
\hline {Dev}   &   {Ses04F}   &   {1105} &   {Train}  &   {2009.05.25.11.23.09} &   {928}  \\
\hline {Dev}   &   {Ses04M}   &   {998} &   {Train}  &   {2009.05.26.10.19.53} &   {790}  \\
\hline {Test}  &   {Ses05F}   &   {1128} &   {Train}  &   {2009.06.05.10.14.28} &   {732}  \\
\hline {Test}  &   {Ses05M}   &   {1042} &   {Train}  &   {2009.06.15.12.13.06} &   {958}  \\
\hline {} &   {}   &   {} &   {Train}  &   {2009.06.19.14.01.24} &   {586}  \\
\hline {} &   {}   &   {} &   {Train}  &   {2009.10.27.16.17.38} &   {440}  \\
\hline {} &   {}   &   {} &   {Dev}  &   {2008.12.19.11.03.11} &   {188}  \\
\hline {} &   {}   &   {} &   {Dev}  &   {2009.01.06.14.53.49} &   {472}  \\
\hline {} &   {}   &   {} &   {Dev}  &   {2009.05.12.15.02.01} &   {448}  \\
\hline {} &   {}   &   {} &   {Dev}  &   {2009.05.08.11.28.48} &   {752}  \\
\hline {} &   {}   &   {} &   {Dev}  &   {2009.06.26.14.38.17} &   {404}  \\
\hline {} &   {}   &   {} &   {Test}  &   {2008.12.14.14.47.07} &   {372}  \\
\hline {} &   {}   &   {} &   {Test}  &   {2009.01.06.12.41.42} &   {264}  \\
\hline {} &   {}   &   {} &   {Test}  &   {2009.01.28.15.35.20} &   {364}  \\
\hline {} &   {}   &   {} &   {Test}  &   {2009.06.26.14.38.17} &   {440}  \\
\hline {} &   {}   &   {} &   {Test}  &   {2009.06.26.14.09.45} &   {448}  \\

\hline
\end{tabular}
\end{table*}

\subsection{Scaling the auxiliary targets}
\label{appendix:scalingaux}
We detail the operations in scaling the percentile scores that range [0,1] to various primary tasks. For IEMOCAP Primary Tasks that are regression problems, we multiply the percentile score by 4 and add 1 to obtain the range [1,5]. For IEMOCAP classification task, we leave the auxiliary targets in the range of [0,1]. As for SEMAINE tasks, which are all regression problems, we multiply the percentile score by 2 and minus 1 to obtain the range [-1,1].

\subsection{Hyperparameters tuning process}
\label{appendix:tuningprocess}
Glove word embeddings (300-dimensions) are used to represent the words \citep{pennington2014glove}. Hyper-parameters tuning is crucial because different combinations of primary and auxiliary tasks require different sets of hyperparameters. For hyperparameters tuning, we used random search \citep{bergstra2012random} with thirty trials. We tuned the learning rate, batch size, L2 regularization, the number of GRUs assigned to the primary and auxiliary branches, the auxiliary weights assignment, the content-context size, and lastly the GRU dropout and recurrent dropout (as detailed in Table \ref{tab:hp_tune_range}).

Training is done on a RTX2070 or a V100 GPU, for up to 350 epochs. Early stopping is possible via the median-stopping rule \citep{golovin2017google} after the fifth epoch and after every two epochs (i.e., at epoch number 5, 7, 9, $...$, 349). Table \ref{tab:besthyparameters} details the hyperparameters of models that performed the best on the development set.

\begin{table*}[ht]
  \centering
  \caption{Range of Hyperparameters tuned. U: Uniform sampling, LU: Uniform sampling on the logarithmic scale.}
  \label{tab:hp_tune_range}  
  \begin{tabular}{|>{\centering\arraybackslash}m{4.00cm}
                |>{\centering\arraybackslash}m{3.10cm}
                |>{\centering\arraybackslash}m{3.10cm}
                |>{\centering\arraybackslash}m{1.35cm}
                |}
\hline {\textbf{Name}}     &   {\textbf{Min.}} &   {\textbf{Max}}   &   {\textbf{Sampling}} \\
\hline {Learning rate}     &   {$2^{-10}$} &   {$2^{-5}$}   &   {LU} \\
\hline {Batch-Size}     &   {32} &   {256}   &   {U} \\
\hline {Pri GRU}     &   \multicolumn{2}{c|}{1, 64, 128, 192, 256}  &   {U} \\
\hline {Aux GRU}     &   \multicolumn{3}{c|}{257 - (minus) Pri GRU}  \\
\hline {Loss-weights Assignment}     &   \multicolumn{2}{c|}{Random, Linear-Normalized, Softmax-Normalized}    &   {U} \\

\hline {L2 Regularization}     &   {0.0} &   {0.50}   &   {LU} \\

\hline {Content-Size}     &   {1} &   {30}   &   {U} \\
\hline {GRU dropout}     &   {0.01} &   {0.50}   &   {U} \\
\hline {Recurrent dropout}     &   {0.01} &   {0.50}   &   {U} \\

\hline
\end{tabular}
\end{table*}

\begin{table*}[ht]
  \centering
  \caption{Best Hyperparameters settings for development set. R: Random, L: Linear-Normalized, S: Softmax-Normalized.}
  \label{tab:besthyparameters}
  \begin{tabular}{|>{\centering\arraybackslash}m{2.40cm}
                |>{\centering\arraybackslash}m{1.00cm}
                |>{\centering\arraybackslash}m{1.00cm}
                |>{\centering\arraybackslash}m{1.00cm}
                |>{\centering\arraybackslash}m{1.00cm}
                |>{\centering\arraybackslash}m{1.00cm}
                |>{\centering\arraybackslash}m{1.00cm}
                |>{\centering\arraybackslash}m{1.00cm}
                |>{\centering\arraybackslash}m{1.00cm}|}
\hline 
        { }     &   \multicolumn{4}{c|}{\textbf{IEMOCAP}} &   \multicolumn{4}{c|}{\textbf{SEMAINE}} \\
\hline
            {\textbf{}}     &   {\textbf{Classif. \newline MA(4)} }
        &   {\textbf{Val. \newline MAE}}   &   {\textbf{Act. \newline MAE}}
        &   {\textbf{Dom. \newline MAE}}  &   {\textbf{Act. \newline MAE}}
        &   {\textbf{Int. \newline MAE}}  &   {\textbf{Pow. \newline MAE}}
        &   {\textbf{Val. \newline MAE}}\\
\hline
            {Dev. set}     &   {0.714} &   {0.482}   &   {0.492}&   {0.570}  
            &   {0.118} &   {0.164}  &   {0.135} &   {0.138}     \\
\hline
            {Test set}     &   {0.747} &   {0.497}   &   {0.499}&   {0.587}  
            &   {0.184} &   {0.215}  &   {0.164} &   {0.169}     \\
\hline
\hline {Learning rate}               &{6.21 e-03} &{2.21 e-02} &{1.44 e-02} &{3.13 e-02}  &{2.43 e-02} &{1.58 e-02} &{3.04 e-03} &{2.94 e-02} \\
\hline {Batch size}                  &{43} &{43} &{62} &{77}  &{41} &{43} &{33} &{62} \\
\hline {Pri GRU}                     &{256} &{256} &{256} &{256}  &{256} &{256} &{192} &{256} \\
\hline {Aux GRU}                     &{1} &{1} &{1} &{1}  &{1} &{1} &{65} &{1} \\
\hline {L2 regularization}           &{2.25 e-05} &{0} &{1.43 e-05} &{2.41 e-04}  &{0} &{0} &{0} &{0} \\
\hline {Content-Size}                &{18} &{4} &{3} &{3}  &{17} &{21} &{19} &{14} \\
\hline {GRU dropout}                 &{0.27} &{0.49} &{0.33} &{0.30}  &{0.06} &{0.17} &{0.05} &{0.07} \\
\hline {Recurrent dropout}           &{0.04} &{0.03} &{0.02} &{0.32}  &{0.25} &{0.28} &{0.08} &{0.26} \\

\hline {Epoch No.}                   &{232} &{161} &{74} &{14}  &{48} &{116} &{110} &{110} \\
\hline {Aux. weights assignment}     &{S} &{R} &{L} &{L}  &{L} &{R} &{S} &{R} \\
\hline {Main loss}                 &{0.840} &{0.560} &{0.810} &{0.610} &{0.870} &{0.940} &{0.950} &{0.920} \\
\hline {AU05 loss}                 &{0.009} &{0.013} &{0.014} &{0.000} &{0.000} &{0.002} &{0.002} &{0.008} \\
\hline {AU17 loss}                 &{0.009} &{0.027} &{0.009} &{0.003} &{0.002} &{0.004} &{0.002} &{0.003} \\
\hline {AU20 loss}                 &{0.009} &{0.028} &{0.0} &{0.001} &{0.000} &{0.007} &{0.002} &{0.006} \\
\hline {AU25 loss}                 &{0.009} &{0.031} &{0.006} &{0.002} &{0.015} &{0.008} &{0.002} &{0.009} \\
\hline {Happy tone loss}           &{0.010} &{0.023} &{0.038} &{0.006} &{0.021} &{0.013} &{0.002} &{0.003} \\
\hline {Sad tone loss}             &{0.010} &{0.028} &{0.053} &{0.008} &{0.037} &{0.010} &{0.002} &{0.003} \\
\hline {Angry tone loss}           &{0.010} &{0.016} &{0.034} &{0.006} &{0.055} &{0.011} &{0.002} &{0.000} \\
\hline {Fear tone loss}            &{0.010} &{0.043} &{0.036} &{0.004} &{0.000} &{0.004} &{0.002} &{0.003} \\
\hline {$Y_{t-1}$ loss}            &{0.010} &{0.030} &{0.0} &{0.040} &{0.000} &{0.000} &{0.005} &{0.006} \\
\hline {$Y_{t-2}$ loss}            &{0.011} &{0.022} &{0.0} &{0.049} &{0.000} &{0.000} &{0.004} &{0.004} \\
\hline {$Y_{t-3}$ loss}            &{0.010} &{0.006} &{0.0} &{0.046} &{0.000} &{0.000} &{0.003} &{0.006} \\
\hline {$Y_{t-4}$ loss}            &{0.010} &{0.040} &{0.0} &{0.044} &{0.000} &{0.000} &{0.003} &{0.007} \\
\hline {$Y_{t+1}$ loss}            &{0.010} &{0.042} &{0.0} &{0.042} &{0.000} &{0.000} &{0.005} &{0.005} \\
\hline {$Y_{t+2}$ loss}            &{0.011} &{0.0} &{0.0} &{0.052} &{0.000} &{0.000} &{0.004} &{0.006} \\
\hline {$Y_{t+3}$ loss}            &{0.010} &{0.050} &{0.0} &{0.043} &{0.000} &{0.000} &{0.003} &{0.008} \\
\hline {$Y_{t+4}$ loss}            &{0.010} &{0.038} &{0.0} &{0.044} &{0.000} &{0.000} &{0.003} &{0.003} \\

\hline
\end{tabular}
\end{table*}

\subsection{Details of computing the bootstrap confidence interval}
\label{appendix:bootstrap}
Baseline models for each hypothesis are detailed in section \ref{section:related_works}. All non-baseline models are referred to as challenger models. We created 1000 bootstrap samples of the test set performance by (1) resampling the development set performance, then (2) selecting the set of hyperparameters that resulted in the best development set performance, and (3) looking up the test set performance given the set of best-performing hyperparameters for the development set. To judge whether the challenger outperforms the baseline, we computed the 95 percent confidence interval by (1) performing element-wise subtraction between the resampled test set performance of the baseline against the challenger, (2) removing the top and bottom 2.5 percent from the differences, and (3) observing whether the remaining 95 percent confidence interval includes zero. If it does not include zero, then the difference is statistically significant.

\subsection{Visualization from HAN-ROCK}
\label{appendix:visualization}

We demonstrate how the HAN-ROCK model could be used to support humans analyze conversations using only text-based inputs. We visualized the attention weights from two models, (1) MTL refers to the classification model with auxiliary supervisors, whilst (2) STL refers to the same model architecture, but its auxiliary supervisors' loss weights are set to zero. In principle, the MTL model should exhibit attention weights that are less likely to overfit because the weights are tempered by auxiliary supervisors. We observe that across the two models, both use the historical talkturns more so than the current talkturn; secondly, both assign high attention to the second word of the talkturns, which is interesting because the second word is where the multimodal annotations are inserted.

\begin{figure*}[ht]
    \includegraphics[width=1.0\textwidth]{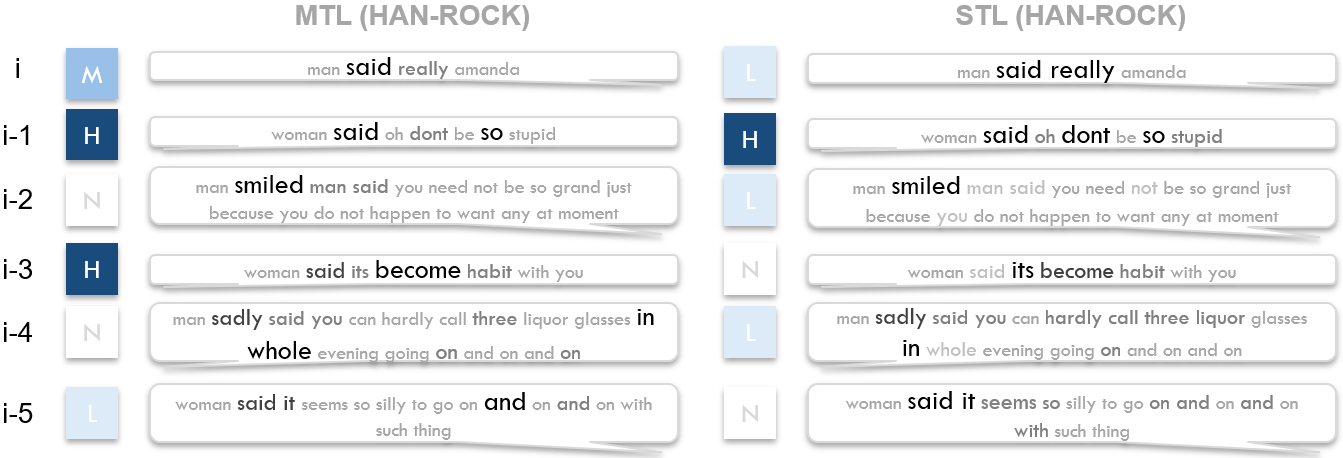}
    \caption{Conversation analysis example. Both models predicted the class label (anger) correctly. The left-most column denotes the talkturn context -- $i$ refers to the current talkturn where the target emotion class is predicted. The square boxes indicate the level of attention (N: None, L: Low, M: Medium, H: High) assigned to the talkturn. Within the talkturn, we also enlarge and darken the font color ti visualize higher attention weights.}
    \label{fig:Visualization}
\end{figure*}

As explained in Section \ref{ssect:han_rock}, there are three levels of attention over the internal representations, word ($\alpha_{it}$), talkturn ($\alpha_{abi}$), and task ($\alpha_{c}$). To compute the overall word and talkturn attention, we compute the weighted average of $\alpha_{it}$ and $\alpha_{abi}$ using the $\alpha_{c}$ (task attention) as weights. Once we have the overall word and talkturn attention, we standardize the weights by computing the z-score. Depending on the z-score, we bucket the attention in none (z $<$ 0), low (0 $<$ z $<$ 1), medium (1 $<$ z $<$ 2), or high (2 $<$ z). We plan to validate the efficacy of the attention weights with human users in future research.

\end{document}